\documentclass[twocolumn,final]{elsarticle}
\usepackage{prletters}
\usepackage[utf8]{inputenc}
\usepackage{PTusefulMacros}
\usepackage{amsmath}
\usepackage{amssymb}
\usepackage[rgb]{xcolor}

\usepackage{hyperref}

\def\ptFiguresDirectory#1{./figures/#1}
\def\ivBrack#1{ {[\![}#1{]\!]} } 
\newsavebox\CBox
\def\textBF#1{\sbox\CBox{#1}\resizebox{\wd\CBox}{\ht\CBox}{\textbf{#1}}}
\def\ptVecX#1#2#3{\left[\ptRowX{#1}{#2}{#3} \right]}

\def\ptRowX#1#2#3{{#1}_{#21},{#1}_{#22},\ldots,{#1}_{#2#3}}

\begin{document}

\begin{frontmatter}
\title{Weighting Scheme for a Pairwise Multi-label Classifier Based on the Fuzzy Confusion Matrix.}


\author[mymainaddress]{Pawel Trajdos\corref{mycorrespondingauthor}}
\cortext[mycorrespondingauthor]{Corresponding author}
\ead{pawel.trajdos@pwr.edu.pl}

\author[mymainaddress]{Marek Kurzynski}

\address[mymainaddress]{Wroclaw University of Science and Technology,  Department of Systems and Computer Networks \\ Wyb. Wyspianskiego 27,  50-370 Wroclaw, Poland}

\begin{abstract}
In this work, we addressed the issue of improving the classification quality of label pairwise ensembles. Our goal is to improve the classification quality achieved by the ensemble via modification of the base classifiers that constitute the ensemble. To achieve this goal, a correction procedure that computes the measures of competence and cross-competence of a single classifier is proposed. These measures are used to modify the prediction of a base classifier. The measures are calculated using a dynamic confusion matrix. Additionally, we provide a weighting scheme that promotes the base classifiers that are the most susceptible to the correction based on the fuzzy confusion matrix. During the experimental study, the proposed approach was compared to two reference methods. The comparison was made in terms of eight different quality criteria. The result shows that the proposed method is able to improve classification quality when compared to baseline methods. 

\end{abstract}


\end{frontmatter}

\section{Introduction}
\label{sect:Intro}
Under  the traditional, supervised classification framework, an object is assigned to only one class. However, many real-world datasets contain objects that are assigned to different categories at the same time. All of these concepts constitute a full description of the object and the omission of one of these tags induces a loss of information. For example, an image may be described using such tags as sea, beach and sunset. The classification process in which such kind of data is involved is called \textit{multi-label} (ML) classification~\cite{Gibaja2014}. In the last 15 years, multi-label learning has been employed in a wide range of practical applications, including text classification~\cite{Jiang2012}, multimedia classification~\cite{Sanden2011} and bioinformatics~\cite{Wu2014}, to name a few.

Multi-label classification algorithms can be broadly partitioned into two main groups i.e. dataset transformation algorithms and algorithm adaptation approaches~\cite{Gibaja2014}.

The method that belongs to the group of algorithm adaptation approaches provides a generalisation of an existing multi-class algorithm. The generalised algorithm is able to solve the multi-label classification problem in a direct way. Among the other methods, the most known approaches from this group are: multi label Nearest Neighbours algorithm~\cite{Jiang2012}, the ML Hoeffding trees~\cite{Read2012}, the structured output support vector machines~\cite{Diez2014} or deep-learning-based algorithms~\cite{Wei2015}.

On the other hand, methods from the former group decompose a multi-label problem into a set of single-label classification tasks. During the inference phase, outputs of the underlying single-label classifiers are combined in order to create a multi-label prediction. An example of this is the \textit{binary relevance} (BR) approach that decomposes a multi-label classification task into a set of \textit{one-vs-rest} binary classification problems~\cite{AlvaresCherman2010}. This approach assumes that labels are conditionally independent, however the assumption does not hold in most real-life recognition problems and the BR framework is considered to be one of the most widespread multi-label classification methods~\cite{Tsoumakas2009}, due to its excellent scalability and acceptable classification quality.

An alternative technique of decomposing the multi-label classification task into a set of binary classifiers, which is studied in depth throughout this article, is the \textit{label-pairwise} (LPW) scheme~\cite{Furnkranz2002}. Under this framework, each pair of labels is assigned with a \textit{one-vs-one} binary classifier. The outcome of the classifier is interpreted as an expression of pairwise preference in a label ranking~\cite{Hllermeier2010}. In contrast to the previously mentioned decomposition technique, the pairwise method considers the paired-inter-label dependencies.  Contrary to the BR approach, this kind of decomposition produces a significantly larger number of base classifiers that must be built. The transformed datasets are, in general, less imbalanced than datasets produced using one-vs-rest transformation. What is more, the resulting base classifiers tend to produce simpler models than one-vs-rest classifiers~\cite{Furnkranz2002}.

This study is conducted with the aim of assessing the results of the application of an \textit{information-theory-based} competence measure when improving the classification quality obtained by label-pairwise multi-label classifiers. Focus is especially put on investigating the impact of the aforementioned quality criterion on a classifier that is corrected using a procedure based on the \textit{fuzzy confusion matrix} (FCM)~\cite{Trajdos2016} and \textit{Random Reference classifier} (RRC)~\cite{Woloszynski2011}. The procedure corrects predictions of the classifiers constituting the LPW ensemble. The outcome of each of the LPW members is individually modified according to the confusion pattern obtained during the validation stage. The members are then combined using a combination method driven by the information-theoretic competence measure.

The concept of the \textit{fuzzy confusion matrix} was first introduced in research aimed at improving the classification quality of systems that recognise hand gestures~\cite{Kurzynski2015,Trajdos2016}. In the above-mentioned research, the FCM-based system was used because it possesses the ability to utilise soft class-assignment. Its ability to improve the response of base classifiers was also seen to be noteworthy. The fuzzy-confusion matrix-based approach was also employed under a multi-label classification framework~\cite{Trajdos2015}. Namely, it was used to improve the quality of Binary Relevance classifiers. Experiments confirmed the validity of its use, but also showed sensitivity to the unbalanced class distribution in a binary problem. This study focuses on addressing this issue via the employment of the LPW technique, which produces more balanced single-label problems than the BR approach. This work is an extension another paper that is yet to be published~\cite{trajdos2017correction}. In the mentioned paper, the application of imbalance reduction procedures on the FCM-based LPW classifier is investigated. In this work, on the other hand, the impact of employing the information-theoretic measure of correction ability is investigated. What is more, in comparison to the previously conducted research, the experimental design has changed significantly in that feature selection and thresholding procedures are employed (For more details see Section~\ref{sect:ExpSet}). 

During the prediction phase, a weight function based on information theory is employed. The decision theoretic measures have proven to be effective methods of assessing the competence of a single-label classifier~\cite{Jurman2012,Wang2013}. The main motivation was that the information-theoretic measures hold a few properties that make them very reliable indicators of the competence of an FCM-corrected classifier. That is, these criteria assess mutual dependence between random variables~\cite{Jurman2012}.

The main motivation of employing an information-theory-based measure was that it allows us to determine if the correction procedure will provide a valuable result. To be more precise, the previously conducted research showed that although the FCM model is able to correct a randomly guessing classifier, the correction is most effective when the underlying base classifier makes a systematic error~\cite{Trajdos2016}. The information-theoretic competence criterion allows such a situation to be detected, and more weight to be put on classifiers with higher correction ability.

This paper is organized as follows. Section~\ref{sect:Proposed} provides the formal notation used throughout the article, and introduces the FCM correction algorithm and its weighted version. Section~\ref{sect:ExpSet} contains a description of the experimental setup. In section~\ref{sect:ResAndDisc}, the experimental results are presented and discussed and section~\ref{sect:Conc} concludes the paper. 

\section{Proposed method}
\label{sect:Proposed}

\subsection{Preliminaries}
\label{subsect:Prelim}
Under the Multi-label formalism, a $d-\mathrm{dimensional}$ object $\vec{x}=\ptVecX{x}{}{d}\in\mathcal{X}=\mathbb{R}^{d}$ is assigned to a set of labels indicated by a binary vector of length $L$: $\vec{y}=\ptVecX{y}{}{L}\in\mathcal{Y}=\{0,1\}^{L}$, where $L$ denotes the number of labels. Each element of the binary vector corresponds to a single label.
In this study, a multi-label classifier $\psi$ is built in a supervised learning procedure using the training set $\mathcal{S}_N$ containing $N$ pairs of feature vectors $\vec{x}$ and corresponding label vectors $\vec{y}$.

The proposed classification method follows the statistical classification framework. As a consequence, vectors $\vec{x}$ and $\vec{y}$ are assumed to be realisations of random variables ${\vec{\textbf{X}}}$ and ${\vec{\textbf{Y}}}$, respectively.

\subsection{Pairwise Transformation}
\label{subsect:PWT}

The label-pairwise (LPW) transformation builds the multi-label classifier $\psi$, using an ensemble of binary classifiers $\Psi$ and a single binary classifier is assigned to each pair of labels:
\begin {equation}   
\Psi=\{\psi_{m},\; m=1,2,...,L(L-1)/2\}.
\end{equation}
During the training phase of a binary classifier $\psi_m$ only learning objects belonging to  either the $m_1$-th or $m_2$-th label are used. Examples that appear in both classes are ignored. Instances assigned to other labels are also ignored because they hold no information that can be used by the binary classifier~\cite{Hllermeier2010}.

During the inference stage, a binary classifier $\psi_{m}(\vec{x})$,at the continuous-valued output level, produces a 2-dimensional vector of label supports $\left[d^{m1}_{m}(\vec{x}), d^{m2}_{m}(\vec{x})\right] \in [0,1]^{2}$. Without loss of generality it is assumed that the output vector is normalised, that is the label specific outputs sum up to one: $d^{m_1}_{m}(\vec{x}) + d^{m_2}_{m}(\vec{x})=1$. Label support $d^{m1}_{m}$ expresses the degree of confidence that the classifier $\psi_{m}(\vec{x})$ gives to the hypothesis that $m1$ is a true class. Higher support stands for better confidence. 

The final decision of the LPW ensemble is obtained by combining the continuous-valued outputs of the binary classifiers that constitute the ensemble. Thus, the final support for the $i$-th label $d^{(i)}$ is calculated using the weighted average of the soft outputs of adequate binary classifiers $d_m^{m_k}(\vec{x})$:
\begin{equation}\label{eq:softRank}
d^{(i)}(\vec{x})= \frac{\sum_{m: m_k = i} w^{m}(\vec{x}) d_m^{m_k}(\vec{x})}{\sum_{m: m_k = i} w^{m}(\vec{x})},
\end{equation}
where $w^{m}(\vec{x})$ is a weight calculated in a dynamic way for the input vector $\vec{x}$ that is assigned to a pair-specific binary classifier. 

The final response of the multi-label classifier $\psi(\vec{x})$ is the binary vector obtained as a result of the thresholding procedure that is applied to the soft outputs, which are calculated using formula~\eqref{eq:softRank}:
\begin{equation}
\label{eq:finalMlClassOutput}
\psi(\vec{x})=\left[ \ivBrack{d^{(1)}(\vec{x})>\theta_1}, \ivBrack{d^{(2)}(\vec{x})>\theta_2},\ldots,\ivBrack{d^{(L)}(\vec{x})>\theta_L}\right],  
\end{equation}
where $\theta_{i}\; i \in \{1,2,\cdots,L\} $ are label-specific thresholds that can be set manually or determined using SCut or RCut procedures~\cite{Yang2001}. 

\subsection{Proposed Correction Method}
\label{subsect:COR}
The proposed correction method is based on an assessment of the probability of classifying an object $\vec{x}$ to the class $h_m \in \{m_1, m_2\}$  using the binary classifier $\psi_m$. The proposed approach provides an extension of the Bayesian model defined in the previous section. Namely, the above-mentioned Bayesian model requires that the object description $\vec{x}$ and its true label $s_m \in \{m_1, m_2\}$ are realizations of random variables  $\vec{\textbf{X}}$ and \textbf{$S_m$}, respectively.  Our method, on the other hand, assumes that the prediction of the classifier $\psi_m$ is made in a random way according to the probabilities $P(\psi_m(\vec{x})=h_m)=P(h_m|\vec{x}) \in [0,1]$~\cite{Berger1985}. As a result the classification result $h_m$ is a realization of the random variable $\mathbf{H}_m$.

The extended Bayesian model allows the posterior probability  $P(s_m|\vec{x})$ of label $s_m$ to be defined as:
\begin{align}\label{eq:postProb1}
 P(s_m|\vec{x}) &= \sum_{h_m \in \{m_1, m_2\}} P(h_m|\vec{x}) P(s_m|h_m,\vec{x}). 
\end{align}
where $P(s_m|h_m,\vec{x})$ denotes the probability that an object $\vec{x}$ belongs to the class $s_m$ given that $\psi_m(\vec{x})=h_m$.

Unfortunately, assuming that the base classifier assigns labels in a stochastic way is rather impractical, because most real-life classifiers are deterministic. This issue was dealt with by employing deterministic binary classifiers in which their statistical properties were modelled using the RRC procedure~\cite{Woloszynski2011}.

\subsection{Confusion Matrix}
\label{subsect:ConfM}

During the inference process of the proposed approach, the probability $P(s_m|h_m,\vec{x})$ is estimated using a local, fuzzy confusion matrix. An example of such a matrix for a binary classification task is given in Table~\ref{MK_PT:confmatrix}. The rows of the matrix correspond to the ground-truth classes, whereas the columns match the outcome of the classifier. The confusion matrix becomes a soft/fuzzy one because the decision regions of the random classifier are expressed in terms of fuzzy set formalism~\cite{Zadeh1965}. Thus, the membership function of a point is proportional to the probability of assigning the point to a given class by the randomised model of the classifier. 

To provide an estimation of $P(s_m|h_m,\vec{x})$ that depends on the description of the instance $\vec{x}$, a confusion matrix that is built using the concept of the neighbourhood of the instance is defined. The neighbourhood of the instance is also defined using the fuzzy set formalism. The fuzzy neighbourhood is employed in order to utilize all the points included in the validation set.

The local fuzzy confusion matrix is estimated using a validation set:
\begin{equation}\label{eq:valSet} 
\mathcal{V}=\left\{(\vec{x}^{1},\vec{y}^{1}), (\vec{x}^{2},\vec{y}^{2}), \ldots ,(\vec{x}^{M},\vec{y}^{M})\right\},
\end{equation}
where $\vec{x}^{k} \in \mathcal{X}$ and $ \vec{y}^{k} \in \mathcal{Y}$ denotes the description of the $k-\mathrm{th}$ instance and the corresponding vector that indicates label assignment, respectively. 
On the basis of this set, pairwise subsets of the validation set, the fuzzy decision region of $\psi_{m}$ and the neighbourhood of $\vec{z}$ were defined, respectively:
\begin{align}
\label{eq:fvs}
  \mathcal{V}^{m}_{s_m} &= \left\{  (\vec{x}^{k},\vec{y}^{k}, 1): (\vec{x}^{k},\vec{y}^{k}) \in \mathcal{V},\right.\nonumber \\
    & \left. y^{k}_{m_1} + y^{k}_{m_2}=1,y^{k}_{s_m}=1  \right\},\\
\label{eq:fds}
  {\mathcal{D}}^{m}_{h_m} &= \left\{  (\vec{x}^{k},\vec{y}^{k} , \mu_{\mathcal{D}^{m}_{h_m}}(\vec{x}^{k}  ) ): (\vec{x}^{k},\vec{y}^{k}) \in \mathcal{V}\right\},\\
\label{eq:fns}
 \mathcal{N}(\vec{z}) &= \left\{  (\vec{x^{k}},\vec{y}^{k} ,\mu_{\mathcal{N}(\vec{z})}(\vec{x^{(k)}}) ):(\vec{x}^{k},\vec{y}^{k}) \in \mathcal{V} \right\},
\end{align}
where each triplet $(\vec{x}^{k},\vec{y}^{k}, \zeta)$ defines the fuzzy membership value $\zeta$ of instance $(\vec{x}^{k},\vec{y}^{k})$, and $\mu_{\mathcal{D}^{m}_{h_m}}(x)=P^{(RRC)}(h_m|\vec{x})$ indicates the fuzzy decision region of the stochastic classifier. Additionally, $\mu_{\mathcal{N}(\vec{z})}(\vec{x})$ denotes the fuzzy neighbourhood of the instance $\vec{z}$. The membership function of the neighbourhood was defined using the Gaussian potential function:
\begin{equation}   \label{MK_PT:mu}
 \mu_{\mathcal{N}(\vec{x})}(z)=\exp({-\beta \delta(\vec{z},\vec{x})^2}),
\end{equation}
where $\beta \in \mathbb{R}_{+}$ and $\delta(\vec{z},\vec{x})$ is a distance  function between two vectors from the input space $\mathcal{X}$.

The above-defined fuzzy sets are employed to approximate $P(s_m|h_m,\vec{x})$:
The following fuzzy sets are employed to approximate entries of the local confusion matrix:
\begin{align}\label{eq:uFCM}
\hat{\varepsilon}^{m}_{s_m,h_m}(\vec{z}) &= \frac{|\mathcal{V}^{m}_{s_m} \cap {\mathcal{D}}^{m}_{h_m} \cap \mathcal{N}(\vec{z})|}{|\mathcal{N}(\vec{z})|}
\end{align}
where $|.|$ is the cardinality of a fuzzy set~\cite{Dhar2013}. Finally, the approximation of $P(s_m|h_m,\vec{x})$ is calculated as follows:
\begin{equation} 
\label{pt:postApprox}
P(s_m|h_m,\vec{x})  \approx \frac{\hat{\varepsilon}^{m}_{s_m,h_m}(\vec{z})}{\sum_{u \in \{ m_1,m_2 \}}\hat{\varepsilon}_{u,h_m}^{m}(\vec{z})}.
\end{equation}


\begin{table}[tb]
\centering\small
\caption{The confusion matrix for a binary classification problem.\label{MK_PT:confmatrix}}
{\begin{tabular}{cc|cc}
& & \multicolumn{2}{c}{estimated}\\
& &  $h_m=m_1$ & $h_m=m_2$\\
\hline
\multirow{2}{*}{true}& $s_m=m_1$& $\varepsilon_{m_1,m_1}^{m}$&$\varepsilon_{m_1,m_2}^{m}$\\
& $s_m=m_2$& $\varepsilon_{m_2,m_1}^{m}$&$\varepsilon_{m_2,m_2}^{m}$\\
\end{tabular}
}
\end{table}

\subsection{Weighting Scheme}
\label{sec:WeightingS}

In this section, the weighting approach that is used during the prediction phase to promote the base classifiers is defined. 

Considering the confusion matrix (Table~\ref{MK_PT:confmatrix}), let us define the boundary distributions within the neighbourhood of $\vec{z}\in\mathcal{X}$, of the true class and classifier response, respectively:
\begin{align} 
\label{eq:MarginalOrg}
f_{s_m}^{m}(\vec{z})  &= \sum_{u \in \{ m_1,m_2 \}}\hat{\varepsilon}_{s_m,u}^{m}(\vec{z}),\\
\label{eq:MarginalClass}
g_{h_m}^{m}(\vec{z})  &= \sum_{u \in \{ m_1,m_2 \}}\hat{\varepsilon}_{u,h_m}^{m}(\vec{z}).
\end{align}
These are used to compute the mutual information and joint entropy of the above-mentioned random variables:
\begin{align}
 \label{eq:MutInfo}
 \mathrm{ICM}^{m}(\vec{z}) &= \sum_{u,v\in \{ m_1, m_2 \}} \frac{\hat{\varepsilon}^{m}_{u,v}(\vec{z})}{f_{u}^{m}(\vec{z}) g_{v}^{m}(\vec{z})}
 \log_{2}\left( \frac{\hat{\varepsilon}^{m}_{u,v}(\vec{z})}{f_{u}^{m}(\vec{z}) g_{v}^{m}(\vec{z})}\right),\\
  \label{eq:ENtropy}
 \mathrm{HCM}^{m}(\vec{z}) &= \sum_{u,v\in \{ m_1, m_2 \}} \hat{\varepsilon}^{m}_{u,v}(\vec{z})\log_{2}\left( \hat{\varepsilon}^{m}_{u,v}(\vec{z})\right).
\end{align}

Finally, the classifier-specific weight is proportional to normalised mutual information~\cite{Cahill2010}:
\begin{equation} 
 \label{eq:inftheoWeight}
 w^{m}(\vec{z}) = \left( \frac{\mathrm{ICM}^{m}(\vec{z})}{\mathrm{HCM}^{m}(\vec{z})} \right)^{\gamma},
\end{equation}
where $\gamma\in(0,1)$ is a factor which decides how uniform the weight is. Thus, for $\gamma$ equals to zero the weights are uniform. For values close to zero, only the classifiers that have a low correction ability are assigned with a low weight. The value of the $\gamma$ coefficient is tuned individually for each classifier during the experimental phase.

\subsection{System Architecture}
\label{subsect:SystemArch}

The detailed description of learning and inference phases are provided in Figures~\ref{pt:table:Train} and~\ref{pt:table:Test}, respectively.
\begin{figure}[tb]
\vbox{%
\begin{center}
\caption{Pseudocode of the learning procedure.}
\label{pt:table:Train}%
\scriptsize
\hrule width 0.45\textwidth
\tt
\smallskip
\begin{tabbing}
\quad \=\quad \=\quad \=\quad \=\quad \=\quad \kill
\textbf{Input data:}\\
\>$\mathcal{S}_N$ - the initial-training set;\\
\>$t \in (0,1)$ -the split percentage;\\
\textbf{BEGIN}\\
\>Split $\mathcal{S}_N$ randomly into $\mathcal{T}$ and $\mathcal{V}$ using $t$\\
\>as the ratio between\\
\>training and validation sets:\\
\>\>$|\mathcal{T}| = t|\mathcal{S}_N|$\\
\>\>$|\mathcal{V}| = (1-t)|\mathcal{S}_N|$\\
\>\>$\mathcal{V},\mathcal{T} \subset \mathcal{S}$, $\mathcal{V} \cap \mathcal{T} = \emptyset$, $\mathcal{T} \cup \mathcal{V} = \mathcal{S}_N$\\
\>Build the LPW ensemble:\\
\>\>$\Psi=\left\{\psi_{m},\; m=1,2,...,L(L-1)/2\right\}$ using $\mathcal{T}$;\\ 
\>For $1\leq m \leq L(L-1)/2$ and $h_m \in \left\{m_1, m_2 \right\}$\\
\>\>build $\mathcal{D}_{h_m}^{m}$ according to \eqref{eq:fds};\\
\>For $1\leq m \leq L(L-1)/2$ and $s_m \in \left\{m_1, m_2 \right\}$\\
\>\>$\mathcal{V}^{m}_{s_m}$ according to \eqref{eq:fvs}\\
\>Save $\mathcal{V}$, $\mathcal{D}_{h}^{m,n}$ and $\mathcal{V}^{m}_{s_m}$;\\
\textbf{END}
\end{tabbing}
\hrule width 0.45\textwidth
\end{center}
}
\end{figure}

\begin{figure}[tb]
\vbox{%
\begin{center}
\caption{Pseudocode of the classification procedure.}%
\label{pt:table:Test}%
\scriptsize
\hrule width 0.45\textwidth
\tt
\smallskip
\begin{tabbing}
\quad \=\quad \=\quad \=\quad \=\quad \=\quad \kill
\textbf{Input data:}\\
\>$\mathcal{V}$ - the validation set;\\
\>$\mathcal{D}_{h_m}^{m}$- decision regions:\\
\>$\mathcal{V}_{s_m}^{m}$- subsets of validation sets:\\
\>$\vec{x}$ - the testing point;\\
\textbf{BEGIN}\\
\>build $\mathcal{N}(\vec{x})$;\\
\>For $1\leq m \leq L(L-1)/2$:\\
\>\> calculate $\hat{\varepsilon}^{m}_{s_m,h_m}(\vec{x})$ according to \eqref{eq:uFCM}\\
\>\>calculate approximations of $P(s_m|h_m,\vec{x})$\\
\>\>according to \eqref{pt:postApprox};\\
\>\>calculate $P(s_m|\vec{x})$ according to \eqref{pt:postApprox};\\
\>\>assign class-specific supports:\\
\>\>\>$d^{m_1}_{m} = P(s_m=m_1|\vec{x}),$\\
\>\>\>$d^{m_2}_{m} = P(s_m=m_2|\vec{x}).$\\
\>\>calculate weight $w^{m}(\vec{x})$ according to~\eqref{eq:inftheoWeight}\\
\>Build final ranking according to \eqref{eq:softRank}\\
\>Convert ranking into\\
\>\>response vector $\psi(\vec{x})$ using \eqref{eq:finalMlClassOutput}\\
\> Return $\psi(\vec{x})$;\\
\textbf{END}
\end{tabbing}
\hrule width 0.45\textwidth
\end{center}
}
\end{figure}

\section{Experimental Setup}
\label{sect:ExpSet}

The conducted experimental study provides an empirical evaluation of the classification quality of the proposed method and compares it to reference methods. Namely, our experiments were conducted using the following algorithms:
\begin{enumerate}
 \item An unmodified LPW classifier~\cite{Hllermeier2010},
 \item An LPW classifier corrected using a confusion matrix.
 \item The LPW classifier corrected using FCM with fusion performed using information theoretic weight.
\end{enumerate}
In the following sections of the paper, the investigated algorithms will be referred to using the above-said numbers.

The following base single-label classifiers were employed: the J48 Tree classifier, which is a weka implementation of C4.5 algorithm~\cite{Quinlan1993}; the Na\"{i}ve Bayes classifier~\cite{Hand2001} combined with attribute selection using the CSF Subset method~\cite{Hall1999}; the voted perceptron algorithm~\cite{Freund1999}.

The thresholds responsible for computing the final prediction vector (see equation~\eqref{eq:finalMlClassOutput}) were computed using the SCut algorithm that was tailored to optimise the macro-averaged $F_1$ loss. 

The $\beta \in \{1,2,\cdots,10 \}$ and $\gamma \in \{2^{-7}, 2^{-6}, \cdots, 2^{-1} \}$ coefficients were determined using grid search and internal 3-CV cross-validation. 

Base classifiers implemented in WEKA framework~\cite{Hall2009} were utilised. The classifier parameters were set to its defaults. All multi-label algorithms were implemented using the MULAN framework~\cite{Tsoumakas2011_mulan}.

The experiments were conducted using 33 multi-label benchmark sets. The main characteristics of the datasets are summarized in Table~\ref{table:Dataset_summ}.

The extraction of training and test datasets was conducted using $10$ fold cross-validation. The proportion of the training set $\mathcal{T}$ was fixed at $t=0.6$ of the original training set $\mathcal{S}_N$ (see Fig.~\ref{pt:table:Train}). Some of the employed sets needed some preprocessing. Thus, multi-label multi-instance~\cite{Zhou2012} sets were used (No.:1,3,4,13,14,24,25), which were transformed to single-instance multi-label datasets according to the suggestion made by Zhou et al.~\cite{Zhou2012}. Two of the used datasets are synthetic ones (No. 27,28) and they were generated using the algorithm described in~\cite{Tomas2014}. To reduce the computational burden, only a subset of original Tmc2007 set was used.

Datasets were used from the sources abbreviated as follows:A --\cite{Charte2015} M--\cite{Tsoumakas2011_mulan}; W--\cite{Wu2014}; X--\cite{Xu2013}; Z--\cite{Zhou2012}; T--\cite{Tomas2014}; S--\cite{DAmbros2010}.

The algorithms were compared in terms of 8 different quality criteria coming from three groups~\cite{Luaces2012}: Ranking-based; Instance-based (Hamming, Zero-One, $F_1$); Label-based. The last group contains the following measures:  Macro Averaged (False Discovery Rate (FDR, 1- Precision), False Negative Rate (FNR, 1-Recall), $F_1$) and Micro Averaged ($F_1$).

Statistical evaluation of the results was performed using the Wilcoxon signed-rank test~\cite{demsar2006,wilcoxon1945} and the family-wise error rates were controlled using the Holm procedure~\cite{demsar2006,holm1979}. For all statistical tests, the significance level was set to $\alpha=0.05$. Additionally, the Friedman~\cite{Friedman1940} test was also applied, followed by the Nemenyi post-hoc procedure~\cite{demsar2006}.

To provide a more detailed look at the properties of the proposed approach, the relations between the classification quality obtained by the investigated algorithms and the chosen dataset characteristics were also analysed. The above-mentioned assessment allows determination of how the investigated classifiers respond to changes in the vital properties of datasets. In order to assess the relations in a quantitative way, the Spearman correlation coefficient was used~\cite{Spearman1904}. The significance of the obtained correlations was tested using two-tailed t-test~\cite{Hollander_2013_book}. As in the experiments related to classification quality, the significance level was also set to $\alpha=0.05$ and the Holm method was employed to adjust p-values~\cite{demsar2006,holm1979}.

{
\setlength\tabcolsep{0.5pt}
\def\arraystretch{0.7}
\begin{table}[tb]
\centering\scriptsize
\caption{Summarised properties of the datasets employed in the experimental study. Sr denotes the source of the dataset, No. is the ordinal number of a set, $N$ is the number of instances, $d$ is the dimensionality of input space and $L$ denotes the number of labels. $\mathrm{LC}$, $\mathrm{LD}$, $\mathrm{avIR}$ and $\mathrm{AVs}$ are label cardinality, label density,  average imbalance ratio and label scumble, respectively~\cite{Luaces2012, Charte2014}. \label{table:Dataset_summ}}
{
\begin{tabular}{l|c|c||cccccccc}
\hline
Name& Sr	& No.&	N&	d&	L&	CD&	LD&	avIR&	AVs\\
\hline
Azotobacter&W&1&407&20&13&1.469&.113&2.225&.010\\
Birds&M&2&645&260&19&1.014&.053&5.407&.033\\
Caenorhabditis&W&3&2512&20&21&2.419&.115&2.347&.010\\
Drosophila&W&4&2605&20&22&2.656&.121&1.744&.004\\
Eclipse-churn&S&5&997&17&5&.264&.053&28.916&.018\\
Eclipse-ent&S&6&997&17&5&.264&.053&28.916&.018\\
Emotions&S&7&593&72&6&1.868&.311&1.478&.011\\
Enron&M&8&1702&1001&53&3.378&.064&73.953&.303\\
Equinox-churn&S&9&324&17&5&.423&.085&41.050&.015\\
Equinox-ent&S&10&324&17&5&.423&.085&41.050&.015\\
Flags&M&11&194&43&7&3.392&.485&2.255&.061\\
Genbase&M&12&662&1186&27&1.252&.046&37.315&.029\\
Geobacter&W&13&379&20&11&1.264&.115&2.750&.014\\
Haloarcula&W&14&304&20&13&1.602&.123&2.419&.016\\
Human&X&15&3106&440&14&1.185&.085&15.289&.020\\
Lucene-churn&S&16&691&17&5&.093&.019&.200&.000\\
Lucene-ent&S&17&691&17&5&.093&.019&.200&.000\\
MimlImg&Z&18&2000&135&5&1.236&.247&1.193&.001\\
Mylyn-churn&S&19&1862&17&5&.236&.047&21.011&.011\\
Mylyn-ent&S&20&1862&17&5&.236&.047&21.011&.011\\
Pde-churn&S&21&1497&17&5&.186&.037&11.454&.012\\
Pde-ent&S&22&1497&17&5&.186&.037&11.454&.012\\
Plant&X&23&978&440&12&1.079&.090&6.690&.006\\
Pyrococcus&W&24&425&20&18&2.136&.119&2.421&.015\\
Saccharomyces&W&25&3509&20&27&2.275&.084&2.077&.005\\
Scene&M&26&2407&294&6&1.074&.179&1.254&.000\\
SimpleHC&T&27&3000&30&10&1.900&.190&1.138&.001\\
SimpleHS&T&28&3000&30&10&2.307&.231&2.622&.050\\
Slashdot&M&29&3782&1079&22&1.181&.054&17.693&.013\\
Stackex\_chemistry&A&30&6961&540&15&1.010&.067&3.981&.024\\
Stackex\_chess&A&31&1675&585&15&1.137&.076&4.744&.025\\
Tmc2007&M&32&2857&500&22&2.222&.101&17.153&.195\\
Yeast&M&33&2417&103&14&4.237&.303&7.197&.104\\
\hline
\end{tabular}}
\end{table}
}

\section{Results and Discussion}
\label{sect:ResAndDisc}

This section shows the results obtained during the conducted experimental study. The following subsections provide a detailed description of the outcome related to classification quality and the dependencies between results obtained by investigated algorithms and the set characteristics, respectively.

\subsection{Classification quality}
\label{ssect:ClassifiQual}
During our study, three different base classifiers were studied for the pairwise ensemble.  As can be seen in Figures~\ref{figure:critRadJ48},~\ref{figure:critRadNB} and~\ref{figure:critRadVPe}, ensembles built using different base classifiers follow the same pattern. Thus, the relation between the investigated multi-label classifiers is quite similar for all base classifiers. 

Due to the page limit in this paper, only the results for the J48 classifier are investigated in a detailed way. The results are shown in Tables~\ref{table:Friedman} and~\ref{table:PW-Hamming}. Full results are presented online~\cite{TrajdosRes2017}.

First of all, it is worth noting that the results reveal that the methods based on FCM correction significantly outperform the unmodified pairwise algorithms in terms of any quality criterion except for Ranking loss. 

What is more, the weighted algorithm outperforms the unweighted FCM approach in terms of the macro averaged $F_{1}$ measure. This result indicates that the proposed weighting scheme allows the FCM classifier to achieve a better performance for rare labels. This phenomenon can be explained by the fact that the weighting scheme assigns lower weights to the FCM classifiers that are biased towards the majority class, since those classifiers cannot be successfully corrected using the FCM approach. As a consequence, the outcome for a given label is produced using base classifiers that were built for more balanced binary sub-problems. The reported property reduces the tendency of the original FCM algorithm to increase the bias towards the majority class~\cite{Trajdos2015} and allows the FCM-based algorithms to be successfully employed in the task of multi-label imbalance classification. This property is also confirmed by the outcomes under macro-averaged FDR and FNR. Thus, the average ranks show that there is almost no difference in terms of the FDR criterion (precision). On the other hand, the average ranks under the FNR criterion suggest that the weighted approach is not so biased towards the majority class as its unweighted counterpart. 

What is more, the classification quality expressed using the micro-averaged $F_{1}$ criterion does not differ significantly between the FCM and its weighted version. It demonstrates that the increase of classification quality for rare labels is not followed by an improvement of classification quality for frequent labels. 

The weighting procedure also causes no significant difference in classification quality under example-based $F_{1}$ loss. Moreover, in the case of micro-averaged and example-based $F_{1}$ measures, the approaches based on the idea of the fuzzy confusion matrix significantly outperform the base label pairwise algorithm. On the other hand, no significant improvement for frequent labels shows that the proposed methods offer almost no improvement when the LPW ensemble is built using label-balanced datasets. However, the base binary classifiers are competent for those datasets. Nonetheless,  these competent classifiers tend to commit systematic errors. As a result, the utilisation of the FCM-based approach allows us to improve classification quality when compared with uncorrected label pairwise ensemble for frequent labels.

The proposed algorithm does not improve the unweighted one in terms of the zero-one quality criterion. The lack of significant improvement under this criterion shows that the proposed method does not achieve a significantly greater number of exact match results among the investigated procedures. When combining these results with the performance achieved under macro-averaged $F_{1}$ loss, it can be concluded that the increase in the perfect match ratio is a consequence of the improved classification of rare labels. However, the increase in the perfect match ratio is not followed by an improvement in terms of zero-one loss. 

The experiments show that assessed classifiers do not differ in a significant way when we consider their ability to produce label ranking instead of a simple binary response. 

The results under the Hamming loss follow the general trends under other loss functions in that the methods based on FCM outperform the base binary-relevance approach. However, the weighting approach does not allow the classification quality to be increased under this criterion. 

{
\setlength\tabcolsep{1.0pt}
\def\arraystretch{0.7}
\begin{table}[tb]
\centering\scriptsize
\caption{P-values for the Friedman test.}
\label{table:Friedman}
\begin{tabular}{r|r||r|r}
  \hline
 Loss & pValue &Loss & pValue\\ 
  \hline
  Hamming & .000048  & Macro FDR & .000005 \\ 
  Zero-One & .000000 & Macro FNR & .000118 \\ 
  EX $F_1$ & .005553 & Macro $F_1$ & .000384 \\ 
  Ranking & .623300 & Micro $F_1$ & .012779 \\ 
  \hline 
\end{tabular} 
\end{table}
}
{
\def\arraystretch{0.8}
\begin{table*}[tb]
\centering\scriptsize
\caption{Wilcoxon test -- p-values for paired comparisons of investigated algorithms. Algorithms are numbered according to Section~\ref{sect:ExpSet}. The last row of the table presents the average ranks achieved over the test sets. \label{table:PW-Hamming}}
\begin{tabular}{c|ccc||ccc||ccc||ccc}
  \hline
  &\multicolumn{3}{c||}{Hamming}	&\multicolumn{3}{c||}{Zero-one}	&\multicolumn{3}{c||}{Ranking}	&\multicolumn{3}{c}{Macro FDR}\\
  \hline
  & 1 & 2 & 3				& 1 & 2 & 3			& 1 & 2 & 3		& 1 & 2 & 3\\ 
  \hline
  1 &  & 0.000 & 0.000			&  & 0.000 & 0.0005		&  & 1.000 & 1.000	&  & 0.000 & 0.000\\ 
  2 &  &  & 0.361			&  &  & 0.148			&  &  & 1.000		&  &  & 0.536\\ 
  \hline
  Rnk & 2.636 & 1.712 & \textBF{1.652}&		2.636 & 1.727 & \textBF{1.636}&		\textBF{1.879} & 2.091 & 2.030	& 2.697 & \textBF{1.636} & 1.667\\
   \hline
   &\multicolumn{3}{c||}{Macro $F_{1}$}		&\multicolumn{3}{c||}{Micro $F_{1}$}		&\multicolumn{3}{c||}{Example $F_{1}$}	&\multicolumn{3}{c}{Macro FNR}\\
  \hline
  1 &  & 0.003 & 0.001				&  & 0.004 & 0.003				&  & 0.007 & 0.005 	&  & 0.000 & 0.000\\ 
  2 &  &  & 0.017				&  &  & 0.057					&  &  & 0.054		&  &  & 0.320\\ 
  \hline
  Rnk & 2.545 & 1.848 & \textBF{1.606}			&2.394 & 1.924 & \textBF{1.682}& 				2.455 & 1.864 & \textBF{1.682}	& \textBF{1.424} & 2.364 & 2.212\\ 
   \hline
   
\end{tabular}
 
\end{table*}
}

{
\begin{figure}[tb]
\begin{center}
   \includegraphics[width=0.33\textwidth]{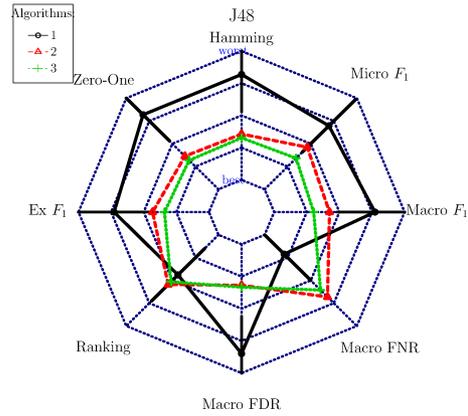}
   \caption{ Base classifier -- J48. Visualisation of the multi-criteria Nemenyi post-hoc test for the investigated algorithms. The black bars parallel to the criterion-specific axes denote the critical difference for the Nemenyi tests.}%
   \label{figure:critRadJ48}
\end{center}
\end{figure}
}

{
\begin{figure}[tb]
\begin{center}
   \includegraphics[width=0.33\textwidth]{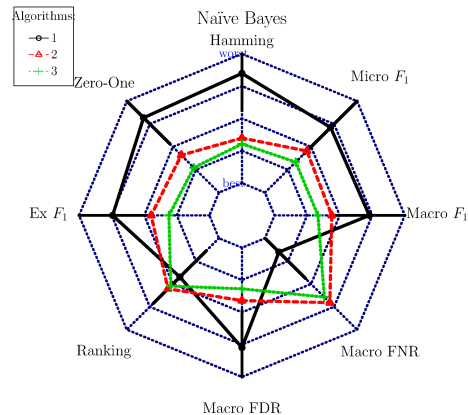}
   \caption{ Base classifier -- Naive Bayes. Radar plot. }%
   \label{figure:critRadNB}
\end{center}
\end{figure}
}

{
\begin{figure}[tb]
\begin{center}
   \includegraphics[width=0.33\textwidth]{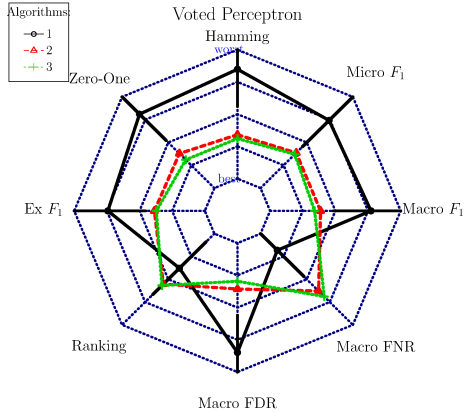}
   \caption{ Base classifier -- Voted Perceptron. Radar plot.}%
   \label{figure:critRadVPe}
\end{center}
\end{figure}
}

\subsection{Impact of dataset properties}
\label{ssect:ResAndDiscSetProperties}

In this section, we assess the relations between the classification quality obtained by a classifier employed on a given multi-label dataset and the properties of this set. At the beginning of the correlation analysis, it is worth mentioning that  the lack of a significant correlation between the multi-label dataset characteristics and the classification quality obtained by an algorithm, under specific circumstances, can be interpreted as an advantage of the classifier.  The algorithm can be seen to be more elastic, as it offers a possibility of being employed in order to solve multi-label classification problems for data sets that significantly differ in terms of characteristics. However, the classifier can only be said to be elastic when it offers an acceptable classification quality for a wide range of data sets. Achieving a satisfactory quality is an important condition, since it is easy to build a classifier that is both completely independent of the set characteristics and that also achieves a low classification quality.

In general, it can be observed in Table~\ref{table:CorRes} that if label density (LD) increases, the classification quality also increases. What is more, in most cases, correlations are significant. This strong correlation is a result of the employment of label pairwise decomposition of the multi-label task. Thus, when LD is high, the instances are better utilised during training and validation phases. In other words, an instance that is relevant to many categories simultaneously more often becomes a member of a training or validation set. As a consequence, the underlying binary classifiers is built using a larger number of training instances. The main exception to this rule is the Ranking loss criterion.

It can also be seen that the classification quality decreases when the imbalance ratio increases. However, this fact is a widely known observation for machine learning~\cite{Lopez2013}, or particularly for under the multi-label classification framework~\cite{Charte2014}. Exceptions to this trend are the results obtained in terms of ranking loss. However, no correlations under this criterion can be considered significant. 

Similarly, no consistent tendency for  the average Scumble measure can be observed. 

{
\def\arraystretch{0.8}
\begin{table*}[tb]
\centering\scriptsize
\caption{Correlations (followed by corresponding p-values) between quality measures and chosen set-specific characteristics. Negative correlations indicate that the classification quality improves (the value of the loss function decreases) when the given characteristic increases. \label{table:CorRes}}
\begin{tabular}{c|ccc|ccc||ccc|ccc}
  \hline
  & \multicolumn{3}{c|}{Hamming} & \multicolumn{3}{c||}{Hamming p-val}	& \multicolumn{3}{c|}{Zero-one} & \multicolumn{3}{c}{Zero-one p-val}\\
  \hline
 & 1 & 2 & 3 	& 1 & 2 & 3	& 1 & 2 & 3 	& 1 & 2 & 3\\ 
  \hline
  LD & -0.668 & -0.423 & -0.370		& 0.000 & 0.113 & 0.240 	& -0.319 & -0.477 & -0.408	& 0.361 & 0.045 & 0.129\\ 
   avIR &  0.335 & 0.187 & 0.140	& 0.339 & 1.000 & 1.000		& 0.309 & 0.331 & 0.306 	& 0.361 & 0.361 & 0.361\\ 
   AVs & -0.151 & -0.129 & -0.127	& 1.000 & 1.000 & 1.000 	& 0.439 & 0.078 & 0.063		& 0.085 & 1.000 & 1.000\\ 
   \hline
   & \multicolumn{3}{c|}{Ranking} & \multicolumn{3}{c||}{Ranking p-val}	& \multicolumn{3}{c|}{Macro $F_{1}$} & \multicolumn{3}{c}{Macro $F_{1}$ p-val}\\
   \hline
   LD & 0.703 & 0.554 & 0.552 		& 0.000 & 0.007 & 0.007		& -0.924 & -0.751 & -0.714		& 0.000 & 0.000 & 0.000\\ 
  avIR & -0.220 & -0.064 & -0.066 	& 0.873 & 1.000 & 1.000		& 0.421 & 0.373 & 0.346 		& 0.088 & 0.162 & 0.195\\ 
  AVs &  0.166 & 0.312 & 0.309		& 1.000 & 0.460 & 0.460		& -0.173 & -0.191 & -0.206		& 0.751 & 0.751 & 0.751\\ 
  \hline
   & \multicolumn{3}{c|}{Micro $F_{1}$} & \multicolumn{3}{c||}{Micro $F_{1}$ p-val}	& \multicolumn{3}{c|}{Example $F_{1}$} & \multicolumn{3}{c}{Example $F_{1}$ p-val}\\
   \hline
  LD & -0.940 & -0.848 & -0.836		& 0.000 & 0.000 & 0.000		& -0.920 & -0.858 & -0.858	& 0.000 & 0.000 & 0.000\\ 
  avIR & 0.369 & 0.390 & 0.377		& 0.153 & 0.148 & 0.153		& 0.347 & 0.345 & 0.345		& 0.289 & 0.289 & 0.289\\ 
  AVs &  0.369 & 0.390 & 0.377		& 1.000 & 1.000 & 1.000		& -0.197 & -0.193 & -0.193	& 0.812 & 0.812 & 0.812\\
  \hline
  & \multicolumn{3}{c|}{Macro FDR} & \multicolumn{3}{c||}{Macro FDR p-val}	& \multicolumn{3}{c|}{Macro FNR} & \multicolumn{3}{c}{Macro FNR p-val}\\
   \hline
  LD & -0.926 & -0.677 & -0.601		& 0.000 & 0.000 & 0.000		& -0.466 & -0.419 & -0.438	& 0.056 & 0.106 & 0.087\\ 
  avIR & 0.409 & 0.332 & 0.281		& 0.108 & 0.295 & 0.454		& 0.288 & 0.164 & 0.175		& 0.626 & 1.000 & 1.000\\ 
  AVs & -0.176 & -0.195 & -0.183	& 0.829 & 0.829 & 0.829		& -0.064 & -0.164 & -0.182	& 1.000 & 1.000 & 1.000\\ 
  
   \hline
\end{tabular} 
\end{table*}
}

\section{Conclusion}
\label{sect:Conc}

During this study, the issue of eliminating the drawbacks of the previously proposed correction algorithm based on the fuzzy confusion matrix was successfully tackled. To reach this goal, we proposed an information theoretic competence measure that assesses if the base binary classifier can take benefits from correction based on the FCM model.

During the experimental study, interesting results were obtained such as the proposed approach is being able to improve classification quality for rare labels (macro-averaged $F_{1}$ loss) and under zero-one loss. What is more, the proposed weighting scheme does not achieve a significantly lower quality in terms of any criterion. In addition, the approach reduces the impact of changing set-specific characteristics. As a consequence, the improved version of the FCM-based algorithm is recommended for use instead of the original one. 

Since the obtained results are promising, we are willing to continue the development of FCM-based algorithms.
\section*{Acknowledgements}
This work is financed by the 'Grant for Young Scientists and PhD Student Development' from the Wroclaw University of Science and Technology, under agreement: 0402/0191/16.
Computational resources were provided by PL-Grid Infrastructure.
   \bibliography{fcmpw}

\begin{thebibliography}{10}
\expandafter\ifx\csname url\endcsname\relax
  \def\url#1{\texttt{#1}}\fi
\expandafter\ifx\csname urlprefix\endcsname\relax\def\urlprefix{URL }\fi
\expandafter\ifx\csname href\endcsname\relax
  \def\href#1#2{#2} \def\path#1{#1}\fi

\bibitem{Gibaja2014}
E.~Gibaja, S.~Ventura, Multi-label learning: {A} review of the state of the art
  and ongoing research, WIREs Data Mining Knowl Discov 4~(6) (2014) 411--444.
\newblock \href {http://dx.doi.org/10.1002/widm.1139}
  {\path{doi:10.1002/widm.1139}}.

\bibitem{Jiang2012}
J.-Y. Jiang, S.-C. Tsai, S.-J. Lee, {FSKNN:} {{Multi}-label} text
  categorization based on fuzzy similarity and k nearest neighbors, Expert
  Systems with Applications 39~(3) (2012) 2813--2821.
\newblock \href {http://dx.doi.org/10.1016/j.eswa.2011.08.141}
  {\path{doi:10.1016/j.eswa.2011.08.141}}.

\bibitem{Sanden2011}
C.~Sanden, J.~Z. Zhang, Enhancing multi-label music genre classification
  through ensemble techniques, in: Proceedings of the 34th international ACM
  SIGIR conference on Research and development in Information - SIGIR '11, ACM
  Press, 2011.
\newblock \href {http://dx.doi.org/10.1145/2009916.2010011}
  {\path{doi:10.1145/2009916.2010011}}.

\bibitem{Wu2014}
J.-S. Wu, S.-J. Huang, Z.-H. Zhou, Genome-wide protein function prediction
  through multi-instance multi-label learning, IEEE/ACM Trans. Comput. Biol.
  and Bioinf. 11~(5) (2014) 891--902.
\newblock \href {http://dx.doi.org/10.1109/tcbb.2014.2323058}
  {\path{doi:10.1109/tcbb.2014.2323058}}.

\bibitem{Read2012}
J.~Read, A.~Bifet, G.~Holmes, B.~Pfahringer, Scalable and efficient multi-label
  classification for evolving data streams, Mach Learn 88~(1-2) (2012)
  243--272.
\newblock \href {http://dx.doi.org/10.1007/s10994-012-5279-6}
  {\path{doi:10.1007/s10994-012-5279-6}}.

\bibitem{Diez2014}
J.~Díez, O.~Luaces, J.~J. del Coz, A.~Bahamonde, Optimizing different loss
  functions in multilabel classifications, Prog Artif Intell 3~(2) (2014)
  107--118.
\newblock \href {http://dx.doi.org/10.1007/s13748-014-0060-7}
  {\path{doi:10.1007/s13748-014-0060-7}}.

\bibitem{Wei2015}
Y.~Wei, W.~Xia, M.~Lin, J.~Huang, B.~Ni, J.~Dong, Y.~Zhao, S.~Yan, {HCP:} {A}
  flexible {CNN} framework for multi-label image classification, IEEE Trans.
  Pattern Anal. Mach. Intell. 38~(9) (2016) 1901--1907.
\newblock \href {http://dx.doi.org/10.1109/tpami.2015.2491929}
  {\path{doi:10.1109/tpami.2015.2491929}}.

\bibitem{AlvaresCherman2010}
E.~Alvares~Cherman, J.~Metz, M.~C. Monard, A simple approach to incorporate
  label dependency in multi-label classification, in: Advances in Soft
  Computing, Springer Berlin Heidelberg, 2010, pp. 33--43.
\newblock \href {http://dx.doi.org/10.1007/978-3-642-16773-7_3}
  {\path{doi:10.1007/978-3-642-16773-7_3}}.

\bibitem{Tsoumakas2009}
G.~Tsoumakas, I.~Katakis, I.~Vlahavas, Mining multi-label data, in: Data Mining
  and Knowledge Discovery Handbook, Springer US, 2009, pp. 667--685.
\newblock \href {http://dx.doi.org/10.1007/978-0-387-09823-4_34}
  {\path{doi:10.1007/978-0-387-09823-4_34}}.

\bibitem{Furnkranz2002}
J.~Fürnkranz, Round robin classification, Journal of Machine Learning Research
  2~(4) (2002) 721--747.

\bibitem{Hllermeier2010}
E.~Hüllermeier, J.~Fürnkranz, On predictive accuracy and risk minimization in
  pairwise label ranking, Journal of Computer and System Sciences 76~(1) (2010)
  49--62.
\newblock \href {http://dx.doi.org/10.1016/j.jcss.2009.05.005}
  {\path{doi:10.1016/j.jcss.2009.05.005}}.

\bibitem{Trajdos2016}
P.~Trajdos, M.~Kurzynski, A dynamic model of classifier competence based on the
  local fuzzy confusion matrix and the random reference classifier,
  International Journal of Applied Mathematics and Computer Science 26~(1).
\newblock \href {http://dx.doi.org/10.1515/amcs-2016-0012}
  {\path{doi:10.1515/amcs-2016-0012}}.

\bibitem{Woloszynski2011}
T.~Woloszynski, M.~Kurzynski, A probabilistic model of classifier competence
  for dynamic ensemble selection, Pattern Recognition 44~(10-11) (2011)
  2656--2668.
\newblock \href {http://dx.doi.org/10.1016/j.patcog.2011.03.020}
  {\path{doi:10.1016/j.patcog.2011.03.020}}.

\bibitem{Kurzynski2015}
M.~Kurzynski, M.~Krysmann, P.~Trajdos, A.~Wolczowski, Multiclassifier system
  with hybrid learning applied to the control of bioprosthetic hand, Computers
  in Biology and Medicine 69 (2016) 286--297.
\newblock \href {http://dx.doi.org/10.1016/j.compbiomed.2015.04.023}
  {\path{doi:10.1016/j.compbiomed.2015.04.023}}.

\bibitem{Trajdos2015}
P.~Trajdos, M.~Kurzynski, An extension of multi-label binary relevance models
  based on randomized reference classifier and local fuzzy confusion matrix,
  in: Intelligent Data Engineering and Automated Learning – IDEAL 2015,
  Springer International Publishing, 2015, pp. 69--76.
\newblock \href {http://dx.doi.org/10.1007/978-3-319-24834-9_9}
  {\path{doi:10.1007/978-3-319-24834-9_9}}.

\bibitem{trajdos2017correction}
P.~Trajdos, M.~Kurzynski, \href{https://arxiv.org/abs/1710.08729}{A correction
  method of a binary classifier applied to multi-label pairwise models}, arXiv
  preprint arXiv:1710.08729\href {http://arxiv.org/abs/1710.08729}
  {\path{arXiv:1710.08729}}.
\newline\urlprefix\url{https://arxiv.org/abs/1710.08729}

\bibitem{Jurman2012}
G.~Jurman, S.~Riccadonna, C.~Furlanello, A comparison of {MCC} and {CEN} error
  measures in multi-class prediction, PLoS ONE 7~(8) (2012) e41882.
\newblock \href {http://dx.doi.org/10.1371/journal.pone.0041882}
  {\path{doi:10.1371/journal.pone.0041882}}.

\bibitem{Wang2013}
X.-N. Wang, J.-M. Wei, H.~Jin, G.~Yu, H.-W. Zhang, Probabilistic confusion
  entropy for evaluating classifiers, Entropy 15~(11) (2013) 4969--4992.
\newblock \href {http://dx.doi.org/10.3390/e15114969}
  {\path{doi:10.3390/e15114969}}.

\bibitem{Yang2001}
Y.~Yang, A study of thresholding strategies for text categorization, in:
  Proceedings of the 24th annual international ACM SIGIR conference on Research
  and development in information retrieval - SIGIR '01, ACM Press, 2001.
\newblock \href {http://dx.doi.org/10.1145/383952.383975}
  {\path{doi:10.1145/383952.383975}}.

\bibitem{Berger1985}
J.~O. Berger, Statistical Decision Theory and {Bayesian} Analysis, Springer New
  York, 1985.
\newblock \href {http://dx.doi.org/10.1007/978-1-4757-4286-2}
  {\path{doi:10.1007/978-1-4757-4286-2}}.

\bibitem{Zadeh1965}
L.~Zadeh, Fuzzy sets, Information and Control 8~(3) (1965) 338--353.
\newblock \href {http://dx.doi.org/10.1016/s0019-9958(65)90241-x}
  {\path{doi:10.1016/s0019-9958(65)90241-x}}.

\bibitem{Dhar2013}
M.~Dhar, On cardinality of fuzzy sets, IJISA 5~(6) (2013) 47--52.
\newblock \href {http://dx.doi.org/10.5815/ijisa.2013.06.06}
  {\path{doi:10.5815/ijisa.2013.06.06}}.

\bibitem{Cahill2010}
N.~D. Cahill, Normalized measures of mutual information with general
  definitions of entropy for multimodal image registration, in: Biomedical
  Image Registration, Springer Berlin Heidelberg, 2010, pp. 258--268.
\newblock \href {http://dx.doi.org/10.1007/978-3-642-14366-3_23}
  {\path{doi:10.1007/978-3-642-14366-3_23}}.

\bibitem{Quinlan1993}
J.~R. Quinlan, {{C4.5} : {Programs} for machine learning}, Morgan Kaufmann
  Publishers Inc., San Francisco, CA, USA, 1993.

\bibitem{Hand2001}
D.~J. Hand, K.~Yu, Idiot's bayes: {Not} so stupid after all?, International
  Statistical Review / Revue Internationale de Statistique 69~(3) (2001) 385.
\newblock \href {http://dx.doi.org/10.2307/1403452}
  {\path{doi:10.2307/1403452}}.

\bibitem{Hall1999}
M.~A. Hall, Correlation-based feature selection for machine learning, Ph.D.
  thesis, The University of Waikato (1999).

\bibitem{Freund1999}
Y.~Freund, R.~E. Schapire, Machine Learning 37~(3) (1999) 277--296.
\newblock \href {http://dx.doi.org/10.1023/a:1007662407062}
  {\path{doi:10.1023/a:1007662407062}}.

\bibitem{Hall2009}
M.~Hall, E.~Frank, G.~Holmes, B.~Pfahringer, P.~Reutemann, I.~H. Witten, The
  {WEKA} data mining software, SIGKDD Explor. Newsl. 11~(1) (2009) 10.
\newblock \href {http://dx.doi.org/10.1145/1656274.1656278}
  {\path{doi:10.1145/1656274.1656278}}.

\bibitem{Tsoumakas2011_mulan}
E.~Spyromitros-Xioufis, G.~Tsoumakas, W.~Groves, I.~Vlahavas, Multi-target
  regression via input space expansion: {Treating} targets as inputs, Mach
  Learn 104~(1) (2016) 55--98.
\newblock \href {http://dx.doi.org/10.1007/s10994-016-5546-z}
  {\path{doi:10.1007/s10994-016-5546-z}}.

\bibitem{Zhou2012}
Z.-H. Zhou, M.-L. Zhang, S.-J. Huang, Y.-F. Li, Multi-instance multi-label
  learning, Artificial Intelligence 176~(1) (2012) 2291--2320.
\newblock \href {http://dx.doi.org/10.1016/j.artint.2011.10.002}
  {\path{doi:10.1016/j.artint.2011.10.002}}.

\bibitem{Tomas2014}
J.~T. Tomás, N.~Spolaôr, E.~A. Cherman, M.~C. Monard, A framework to generate
  synthetic multi-label datasets, Electronic Notes in Theoretical Computer
  Science 302 (2014) 155--176.
\newblock \href {http://dx.doi.org/10.1016/j.entcs.2014.01.025}
  {\path{doi:10.1016/j.entcs.2014.01.025}}.

\bibitem{Charte2015}
F.~Charte, A.~J. Rivera, M.~J. del Jesus, F.~Herrera, {QUINTA:} {A} question
  tagging assistant to improve the answering ratio in electronic forums, in:
  IEEE EUROCON 2015 - International Conference on Computer as a Tool (EUROCON),
  IEEE, 2015.
\newblock \href {http://dx.doi.org/10.1109/eurocon.2015.7313677}
  {\path{doi:10.1109/eurocon.2015.7313677}}.

\bibitem{Xu2013}
J.~Xu, Fast multi-label core vector machine, Pattern Recognition 46~(3) (2013)
  885--898.
\newblock \href {http://dx.doi.org/10.1016/j.patcog.2012.09.003}
  {\path{doi:10.1016/j.patcog.2012.09.003}}.

\bibitem{DAmbros2010}
M.~D'Ambros, M.~Lanza, R.~Robbes, An extensive comparison of bug prediction
  approaches, in: 2010 7th IEEE Working Conference on Mining Software
  Repositories (MSR 2010), IEEE, 2010.
\newblock \href {http://dx.doi.org/10.1109/msr.2010.5463279}
  {\path{doi:10.1109/msr.2010.5463279}}.

\bibitem{Luaces2012}
O.~Luaces, J.~Díez, J.~Barranquero, J.~J. del Coz, A.~Bahamonde, Binary
  relevance efficacy for multilabel classification, Prog Artif Intell 1~(4)
  (2012) 303--313.
\newblock \href {http://dx.doi.org/10.1007/s13748-012-0030-x}
  {\path{doi:10.1007/s13748-012-0030-x}}.

\bibitem{demsar2006}
J.~Dem{\v{s}}ar, Statistical comparisons of classifiers over multiple data
  sets, The Journal of Machine Learning Research 7 (2006) 1--30.

\bibitem{wilcoxon1945}
F.~Wilcoxon, Individual comparisons by ranking methods, Biometrics Bulletin
  1~(6) (1945) 80.
\newblock \href {http://dx.doi.org/10.2307/3001968}
  {\path{doi:10.2307/3001968}}.

\bibitem{holm1979}
S.~Holm, {{A} Simple Sequentially Rejective Multiple Test Procedure},
  Scandinavian Journal of Statistics 6~(2) (1979) 65--70.
\newblock \href {http://dx.doi.org/10.2307/4615733}
  {\path{doi:10.2307/4615733}}.

\bibitem{Friedman1940}
M.~Friedman, A comparison of alternative tests of significance for the problem
  of $m$ rankings, Ann. Math. Statist. 11~(1) (1940) 86--92.
\newblock \href {http://dx.doi.org/10.1214/aoms/1177731944}
  {\path{doi:10.1214/aoms/1177731944}}.

\bibitem{Spearman1904}
C.~Spearman, The proof and measurement of association between two things, The
  American Journal of Psychology 15~(1) (1904) 72.
\newblock \href {http://dx.doi.org/10.2307/1412159}
  {\path{doi:10.2307/1412159}}.

\bibitem{Hollander_2013_book}
M.~Hollander, D.~A.~Wolfe, E.~Chicken, Nonparametric Statistical Methods, John
  Wiley \& Sons, Inc., 2015.
\newblock \href {http://dx.doi.org/10.1002/9781119196037}
  {\path{doi:10.1002/9781119196037}}.

\bibitem{Charte2014}
F.~Charte, A.~Rivera, M.~J. del Jesus, F.~Herrera, Concurrence among imbalanced
  labels and its influence on multilabel resampling algorithms, in: Lecture
  Notes in Computer Science, Springer International Publishing, 2014, pp.
  110--121.
\newblock \href {http://dx.doi.org/10.1007/978-3-319-07617-1_10}
  {\path{doi:10.1007/978-3-319-07617-1_10}}.

\bibitem{TrajdosRes2017}
P.~Trajdos, Results for paper: {Weighting} scheme for a pairwise multi-label
  classifier based on the fuzzy confusion matrix.,
  \url{https://github.com/ptrajdos/MLResults/tree/master/FCMPW_W} (2017).

\bibitem{Lopez2013}
V.~López, A.~Fernández, S.~García, V.~Palade, F.~Herrera, An insight into
  classification with imbalanced data: {Empirical} results and current trends
  on using data intrinsic characteristics, Information Sciences 250 (2013)
  113--141.
\newblock \href {http://dx.doi.org/10.1016/j.ins.2013.07.007}
  {\path{doi:10.1016/j.ins.2013.07.007}}.

\end{thebibliography}
\end{document}